\pdfoutput=1

\documentclass[11pt]{article}

\usepackage[]{ACL2023}

\usepackage{times}
\usepackage{latexsym}

\usepackage[T1]{fontenc}

\usepackage[utf8]{inputenc}

\usepackage{microtype}

\usepackage{inconsolata}
\usepackage{dblfloatfix}

\definecolor{allhard}{rgb}{0.554, 0.180, 0.855}
\definecolor{allinvhard}{rgb}{0.855, 0.180, 0.820}

\usepackage{xcolor}
\usepackage{booktabs}
\usepackage{graphicx}

\newcommand{\fact}[3]{\textit{(#1, #2, #3)}}
\newcommand{\mo}{{\sc Eredat}}
\newcommand{\wchunks}{{\sc WikiChunks}}
\newcommand{\wnlg}{{\sc WebNLG}}
\newcommand{\wnlgwd}{{\sc WebNLG-WD}}
\newcommand{\wnlginv}{{\sc WebNLG-INV}}
\newcommand{\wnlgdb}{{\sc WebNLG-DB}}

\newcommand{\trex}{{\sc TREx}}
\newcommand{\kelm}{{\sc KELM}}
\newcommand{\TeKGen}{{\sc TeKGen}}

\title{Joint Representations of Text and Knowledge Graphs for Retrieval and Evaluation}

\author{Teven Le Scao \\
  Université de Lorraine \\
  \texttt{teven.le-scao@loria.fr} \\\And
  Claire Gardent \\
  Université de Lorraine \\
  \texttt{claire.gardent@loria.fr} \\}

\begin{document}

\maketitle

\begin{abstract}
  A key feature of neural models is that they can produce semantic vector representations of objects (texts, images, speech, etc.) ensuring that similar objects are close to each other in the vector space. While much work has focused on learning representations for other modalities, there are no aligned cross-modal representations for text and knowledge base (KB) elements. One challenge for learning such representations is the lack of parallel data, which we use contrastive training on heuristics-based datasets and data augmentation to overcome, training embedding models on (KB graph, text) pairs. On \wnlg{}, a cleaner manually crafted dataset, we show that they learn aligned representations suitable for retrieval. We then fine-tune on annotated data to create \mo{} (Ensembled Representations for Evaluation of DAta-to-Text), a similarity metric between English text and KB graphs. \mo{} outperforms or matches state-of-the-art metrics in terms of correlation with human judgments on \wnlg{} even though, unlike them, it does not require a reference text to compare against.
\end{abstract}

\section{Introduction}

Neural approaches have progressed in capturing semantic relatedness between larger and larger text units, from Word2Vec \citep{word2vec} to SBERT \cite{reimers-gurevych-2019-sentence}. Such models have shown to perform well on a wide array of semantic similarity tasks, helped in part by retrieval systems like DPR \citep{karpukhin-etal-2020-dense}. 

Other work has shown that  deep representations of knowledge bases (KBs) help improve such tasks as few shot link prediction, analogical reasoning \cite{pezeshkpour-etal-2018-embedding,pahuja-etal-2021-systematic}, entity linking \cite{yu-etal-2020-bridging} or cross-lingual entity alignment \cite{ijcai2018-556,xu-etal-2019-cross-lingual}.

In this work, we focus on learning cross-modal representations for English text and KB graphs. Our input graphs are in RDF (Resource Description Framework, \cite{miller1998introduction}) format, a standard where graphs are sets of \fact{subject}{predicate}{object} triples. We linearize those graphs and consider them as text data so that the same model can take text and graphs as input. Given some aligned RDF-text data, our model learns fixed-length latent representations for texts and RDF graphs such that texts and RDF graphs that are semantically similar are close in vector space. This enables retrieval across modalities and allows us to create a cross-modality similarity score which can be used to evaluate the output of RDF-to-text generation models.

One challenge for learning cross-modal RDF-text representations is the lack of parallel data. We train on various RDF-text datasets created using distant supervision techniques, either combining these datasets or using them in isolation. We then compare the performance of the resulting retrieval models (i) on the \wnlg\ dataset, a parallel RDF-text dataset where texts are crowdsourced to match the graph (texts and graphs are semantically equivalent), and (ii) on \wchunks{}, a more challenging, less well aligned dataset which imitates the conditions in which retrieval on Wikipedia is usually executed. We use the difference in performance between models to analyze the alignment quality of training datasets.

Distance within embedding space can be used to evaluate the output of RDF-to-text generation models (Is the generated text similar to the input graph?). In order to evaluate this metric, we compute correlations between our model's similarity score for graph-text pairs and human judgments of semantic adequacy (input/output semantic similarity) using ratings from the 2020 \wnlg{} Challenge. After fine-tuning on data from the 2017 \wnlg{} challenge, as well as introducing new classes of data augmentation at pre-training time, our best system, \mo{}, is better or on par than existing metrics at correlating with human evaluation, even though it does not require a reference for comparison as do most NLG evaluation metrics such as BLEU \citep{papineni-etal-2002-bleu}, TER \citep{snover-etal-2006-study}, BLEURT \citep{sellam-etal-2020-learning}, METEOR \citep{banerjee-lavie-2005-meteor} or BERT-Score \citep{bert-score}.

Our contributions can be summarised as follows.

\begin{itemize}
\item
We train a cross-modal RDF-text model to learn aligned (RDF graph, text) representations, making it suitable for cross-modal retrieval. We show that this retrieval model outperforms a state-of-the-art text-only retrieval model by a large margin, demonstrating the effectiveness of our adaptation procedure. We train on several datasets of RDF-text pairs, using the quality of the ensuing retrieval models to analyze the quality of training datasets.
\item
We provide a novel evaluation metric for RDF-to-text generation models by combining bi- and cross-encoder training procedures and adding adversarial data to address the models' weaknesses. We show that this new metric outperforms other existing RDF-to-text evaluation metrics in terms of correlation with human judgments of semantic adequacy, even though it does not require a costly human reference to compare against. 
\end{itemize}

\section{Related Work}
\label{sec:related}

We briefly review recent approaches to uni- and cross-modal retrieval, representation learning models, and evaluation metrics for Natural Language Generation (NLG) models.

\paragraph{Natural Language Retrieval Models.} 
For natural language, a first class of retrieval models focuses on retrieving sentences that are similar to some input sentence. BERT \citep{devlin-etal-2019-bert} has been used as a cross-encoder. Two sentences are given with a separator token, cross-attention applies to all input tokens and the resulting representation is fed into a linear layer to score the match. However, this is computationally inefficient as it is not possible to pre-compute and index such representations. A  pre-computable model was proposed by \citep{reimers-gurevych-2019-sentence} who used twin encoders pre-trained on Natural Language Inference data  \citep{bowman-etal-2015-large} to set new state-of-the-art performance on a large set of sentence scoring tasks. Further work \citep{chen-etal-2020-dipair,Polyencoders} combined cross- and bi-encoders to reach a tradeoff between accuracy and efficiency. We differ from those works in that we focus on cross-modal representation learning.

\paragraph{Representation Learning for Knowledge-Bases.} Various KB embedding models have been proposed to support downstream applications such as KB completion or alignment of different bases. Compositional approaches \cite{10.5555/3104482.3104584,10.5555/3016100.3016172} use tensor products to model relations as functions of their argument entities. Translational approaches model relations as translation operations from the subject (head) to object (tail) entity \cite{10.5555/2999792.2999923,yang2014,10.5555/3045390.3045609}. Neural models have also leveraged 2-D convolutions over entity embeddings to predict relations \cite{dettmers2018convolutional} as well as graph convolutional networks \cite{inbook}. All these approaches focus on representation learning for Knowledge-Bases entities and relations. In contrast, we focus on cross-modal similarity between a text and a KB graph. 

\paragraph{Cross-Modal Representation Learning and Retrieval.} Some work has focused on incorporating natural language information to improve KB representations. \cite{Han2016JointRL,toutanova-etal-2015-representing,wu2016knowledge} encode words and KB entities into a single vector space, and \cite{10.5555/3060621.3060801,yamada-etal-2016-joint} learn word and entity embeddings separately then map them into a shared space. Both approaches use text as additional training signal to improve KB representations, and limit themselves to word-level information. Instead, we focus on scoring the similarity between arbitrary-length natural language text and a KB graph. We are not aware of any extant such text-KB models. The best-known cross-modal contrastive model is \citet{Radford2021LearningTV}, which pre-trained an image-text match scoring model.

\paragraph{Evaluation metrics for Natural Language Generation Models.} Surface-based metrics such as BLEU \citep{papineni-etal-2002-bleu}, which measure token overlap between generated and reference text, are commonly used. Methods such as BERT-Score \citep{bert-score} or BLEURT \citep{sellam-etal-2020-bleurt} which leverage neural representations are currently state-of-the-art. All these methods compute a score by comparing the generated text with human-produced references, rarely available and costly to produce. 
Some metrics evaluate the generated output with respect to the input rather than to a reference. \citet{wiseman-etal-2017-challenges} use the precision of input relations found in the output texts. \cite{dusek-kasner-2020-evaluating} use a natural language inference pre-trained model to score input-output two-way entailment. For data-to-text generation specifically, \cite{rebuffel-etal-2021-data} introduce Data-QuestEval, which uses question answering to compare input graph and output text.

\section{Learning Cross-Modal RDF-text Representations}
\label{sec:method}

\subsection{Model}
\label{subsec:model}

Similar to \cite{facenet2015,reimers-gurevych-2019-sentence}, we use twin Transformer encoders to create RDF and text representations such that the embeddings of an RDF graph and of a piece of text with similar content are close in the vector space. A mean-pooling operation creates fixed-sized embeddings $embed(x)$ for $x$ either an RDF graph or a text. RDF graphs are linearized as:

\vskip 1em

\texttt{[S] <subject\textsubscript{1}> [P] <property\textsubscript{1}> [O] <object\textsubscript{1}> ... [S] <subject\textsubscript{n}> [P] <property\textsubscript{n}> [O] <object\textsubscript{n}>} 

\vskip 1em

where "[S]", "[P]", "[O]" serve as special tokens and are added to the tokenizer vocabulary. This allows us to treat any knowledge base format.

We train this system using a contrastive loss with \textit{in-batch negatives} \citep{InBatchNegatives}. This variant of contrastive loss computes the pairwise similarities between every text and every RDF in the batch. A softmax is then applied on the RDF axis, which creates a multi-class classification problem: every text data point must be matched to the parallel RDF. The loss can be written as :

\[
l = -\sum_{i \in I} \log \left( \frac{\exp(sim(text_i, rdf_i))}{\sum_{j \in J} \exp(sim(text_i, rdf_j))} \right)
\]
\[ 
sim(text_i, rdf_j) = \cos(embed(text_i), embed(rdf_j))
\]

with $I$ the set of training instances in the batch. 
Intuitively, this trains the encoder to learn representations that map text items closer to their RDF anchor than to other RDF graphs in the dataset.

In all our experiments, we start from \texttt{all-mpnet-base-v2}, a pre-trained sentence-MPNet \citep{MPNet} model, in order to leverage its strong pre-trained text representations.

\subsection{Training Datasets}

For training, we need $(g,t)$ pairs where $g$ is a Wikidata RDF graph and $t$ is a text in English whose content is similar to $g$. We compare three datasets, all created using distant supervision.

\paragraph{TeKGen.} \cite{agarwal-etal-2021-knowledge} use heuristics to align triples from Wikidata to Wikipedia sentences. The \TeKGen\ dataset covers 1,041 Wikidata properties and consists of about 6M (graph, text) pairs where each text is a sentence.

\paragraph{KELM.} The \kelm{} corpus has 15M (graph, text) pairs where graphs are created based on relation co-occurrence counts i.e. frequency of alignment of two properties to the same sentence in the training data \citep{agarwal-etal-2021-knowledge}. Texts are then generated from these graphs using a T5 model fine-tuned on \TeKGen{}.%

\paragraph{TREx.} 
\cite{elsahar-etal-2018-rex} use word- and sentence-tokenization, coreference resolution, a date-time and a predicate linker, plus various RDF-text alignment methods to create \trex{}, a dataset aligning 11 million Wikidata triples with 6 million Wikipedia sentences.


\subsection{Test Datasets}
We use two datasets for evaluation: \wnlg\ \cite{gardent-etal-2017-creating} and \wchunks, which we create in this work. Appendix~\ref{sec:retrievaldata} shows some statistics for all datasets.

\textbf{WebNLG} is a dataset of pairs where the texts were crowdsourced to match the input graph. In \wnlg\ the RDF graph is from the DBpedia KB, whereas our models were trained on the Wikidata KB format. To assess the ability of our retrieval model to generalize to different KBs, we evaluate our model both on \wnlgdb, the original DBpedia-based dataset, and \wnlgwd\ where the DBpedia graphs have been mapped to Wikidata  \cite{hangenerating}.

\begin{figure*}[ht]
    \centering
    \includegraphics[width=0.9\linewidth]{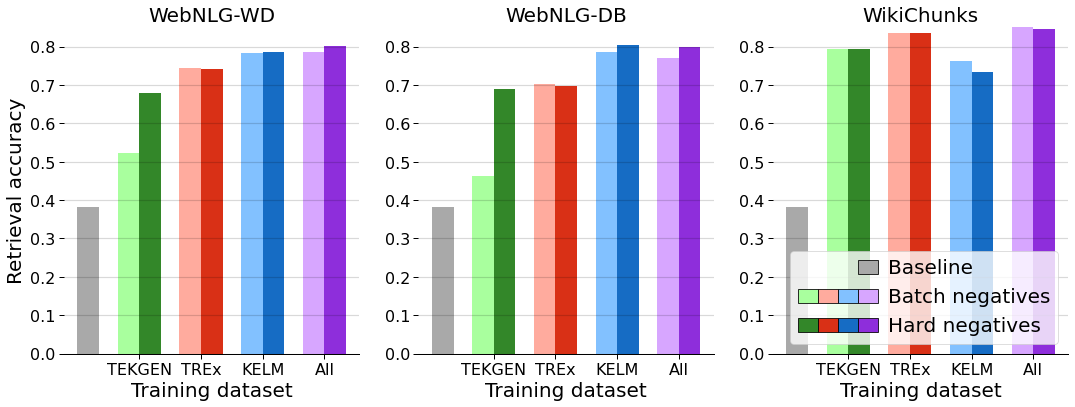}
    \caption{\textbf{Retrieval Accuracy} for a variety of training datasets and objectives. Our models outperform the base model (leftmost grey bar) by a large margin. Hard negatives help across the board. Training on an equal mix of datasets yields consistently high performance on aligned (\wnlg{}) and noisy (\wchunks{}) data.}.
    \label{fig:hardinbatch}
\end{figure*}

\textbf{WikiChunks} 
consists of 7.3M graph-text pairs where the text is a 100-word \textit{passage} from a Wikipedia dump and the graphs are matching Wikidata graphs. We create matching graphs by aligning all  Wikidata  \fact{s}{p}{o} triples with a Wikipedia passage such that the subject $s$ of that triple matches the entity described by the Wikipedia page from which the passage was extracted and the object  $o$, or one of its aliases, is mentioned in that passage. Retrieving on this dataset imitates the conditions in which retrieval on Wikipedia is usually executed \citep{DPR, DPR-RAG}. This is a challenging task as, contrary to \wnlg, \wchunks\ matches are not aligned: the Wikidata graph information is strictly included in the passage, which may contain much more. Several passages may also contain very similar information. We use a subset of 30000 pairs, the same size as \wnlg{}, to make results comparable.

We evaluate our representations using a retrieval reformulation of the data-to-text NLG task: Given the embedding of a graph, how well can we identify the most similar text in the corpus? As our evaluation sets have 1-to-1 mappings between sources (the graphs) and targets (the texts), the retrieval performance in the opposite direction does not vary by more than 2\%. We consider top-result accuracy. 

\section{Results}

\subsection{General Results}

We use \texttt{all-mpnet-base-v2}, the state-of-the-art dense sentence embedding model that our models are training from, as a baseline. \texttt{all-mpnet-base-v2} can estimate semantic similarity, as our models do, but was only trained on text. It can still process the linearized RDF data, however, as it is in the form of natural text. \textbf{The baseline is reasonable, but training yields strong improvements} with a top accuracy of 80\% for all settings against 38\% for the base model (Figure \ref{fig:hardinbatch}) and 0.003\% for random-chance performance. 

\subsection{Generalization to other KB formats}

Encoding the RDF data as natural language allows for flexibility in the RDF format, as opposed to earlier graph approaches that encode relations and entities as integers. After fine-tuning on Wikidata graphs, which include relations like \texttt{place served by transport hub}, we might be able to generalize to DBPedia, which would use \texttt{cityServed} instead, as the base pre-trained model knows all these words. Indeed, we find that \textbf{retrieval performance is similar on \wnlgwd{} and \wnlgdb{}}.

\subsection{Batch Size and Negatives}

We experiment with adding artificial hard negatives to the batch, and with different batch sizes. Confounders are constructed from the correct graph by corrupting a triple inside that graph, replacing a subject, object or predicate at random with another subject, object or predicate in the dataset. This form of data augmentation is made possible by the formalized nature of RDF graphs: it would be much harder to create confounders on the text side.

\paragraph{Hard vs. In-batch negatives} Figure \ref{fig:hardinbatch} shows retrieval accuracy when using only in-batch vs. using in-batch and hard negatives. We see that hard negatives mostly help when retrieving parallel data (\wnlg{}) i.e. when small graph-text mismatches strongly impact accuracy. We also see that hard negatives have the strongest impact on the model trained on \TeKGen{}, which is also the one with the lowest retrieval accuracy. This suggests that hard negatives are most helpful when the training data is noisier than the evaluation data. 

\begin{figure*}[!ht]
    \centering
    \includegraphics[width=0.9\linewidth]{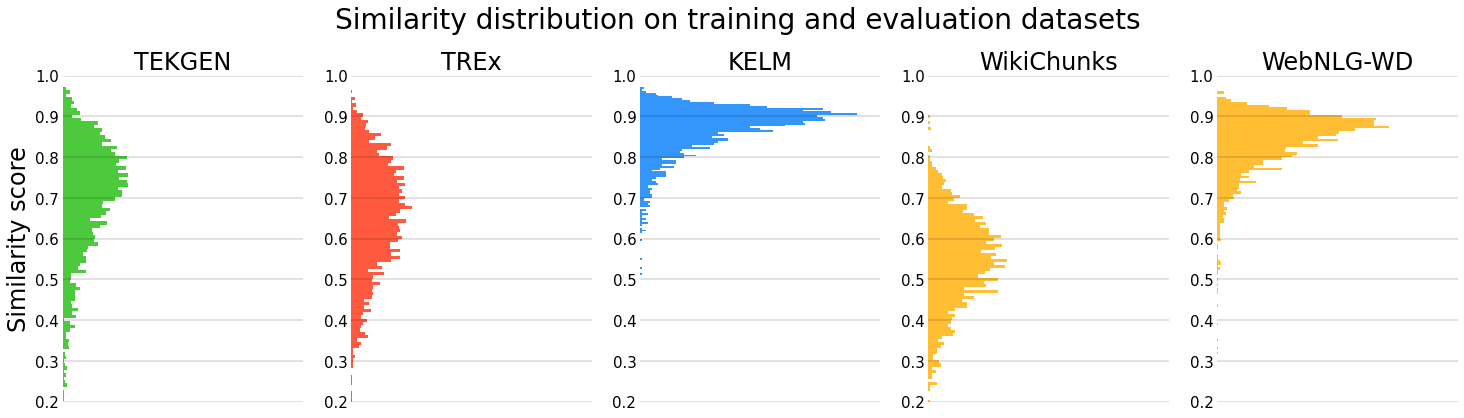}
    \caption{{\bf Pair similarity distributions} according to \textcolor{allhard}{\texttt{all\_datasets\_hard\_negatives}} for all datasets.}
    \label{fig:simdists}
\end{figure*}

\begin{figure*}[!ht]
    \centering
    \includegraphics[width=0.9\linewidth]{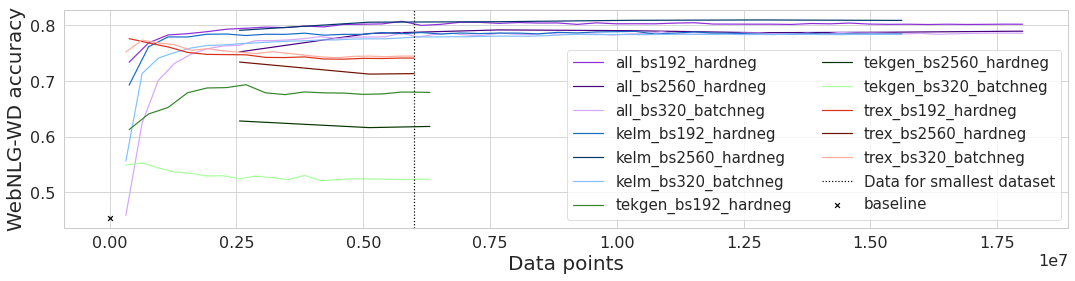}
    \caption{\textbf{Performance throughout training} evaluated by \wnlgwd{} accuracy. Training for longer than the size of the smallest datasets does not change performance meaningfully.}
    \label{fig:performancethroughouttraining}
\end{figure*}

\paragraph{Batch size.} As previous work has found that larger batch sizes improve contrastive training \citep{RocketQA}, we experiment with two batch size set-ups: 192\footnote{The maximum we could fit on an 8-A100 cloud instance.} and 2560\footnote{The maximum we could fit on a larger cluster.}. We do not find that larger batch sizes consistently improve retrieval accuracy, and keep the smaller ones for practical reasons. Figure \ref{fig:batchsize} in appendix \ref{sec:batchsize} shows detailed results.

\subsection{Training Data Quality} 

The quality of training data has a strong impact on retrieval accuracy. We see that performance varies with the training data used: on \wnlg{} retrieval, \kelm{} yields by far the best results followed successively by \trex{} and \TeKGen{}. On \wchunks, which is more loosely aligned, \trex{} is the best dataset and \kelm{} is slightly behind. We create an equal-mixture dataset by concatenating subsets of equal sizes of each dataset\footnote{In total, thrice the size of the smallest dataset, \trex{}.}. As the rightmost column in Figure \ref{fig:hardinbatch} shows, this allows us to capture the best of both worlds. We dub the model trained on this data with hard negatives \textcolor{allhard}{\texttt{all\_datasets\_hard\_negatives}}.

The similarity distributions according to \textcolor{allhard}{\texttt{all\_datasets\_hard\_negatives}} is shown in Figure~\ref{fig:simdists}, which matches those results: \kelm{} is much better aligned. This is in line with intuition as \kelm{} text is generated from the input graphs while \trex{} and \TeKGen{} are created using distant supervision. We attempted to bootstrap dataset quality by re-training models on the 50\% of the data identified as highest-similarity. We find that this does not increase performance and can even decrease it, probably due to loss of diversity.

\subsection{Training Data Quantity}

As shown in Figure \ref{fig:performancethroughouttraining}, performance plateaus early in training. The advantage of \kelm{} or the concatenated dataset is not due to their larger size.

\section{Building a Referenceless Metric for Data-to-text Generation}

Commonly-used metrics for Natural Language Generation require references to compare the output against, which must be produced by human annotators. Can we leverage our joint embeddings to compare the output text to the input RDF directly, reducing the necessary resources?

\begin{figure*}[!ht]
    \centering
    \includegraphics[width=0.78\linewidth]{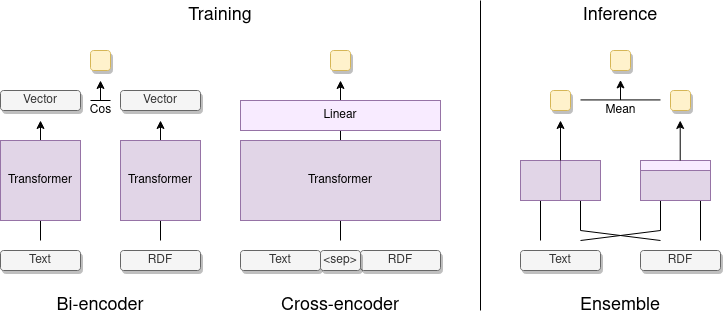}
    \caption{\textbf{Fine-tuning setup.} We fine-tune both bi-encoders and cross-encoders on human-rated data. At inference time, we use the mean of a bi-encoder and a cross-encoder as the final metric.}
    \label{fig:ensembling_archs} 
\end{figure*}

\subsection{Fine-tuning on Human Judgments of Semantic Adequacy}
Our retrieval models can be used to provide a similarity metric between text and formal data in the form of the scalar product or cosine distance in embedding space. We can further improve this metric by fine-tuning on human judgments of RDF-text adequacy. In order to show the generalization strength of this approach, we fine-tune our \textcolor{allhard}{\texttt{all\_datasets\_hard\_negatives}} model on human-rated \wnlg{}-2017 items, and evaluate on human-rated \wnlg{}-2020 items, which uses different test data and different criteria for the assessment of semantic adequacy by human judges.

\cite{webnlg2017human} provides human judgments for the output of 10 NLG systems from \wnlg{} challenge 2017. Each model was evaluated on a sample of 223 texts yielding a total of 2230 generated texts annotated with human judgments for the following three criteria.

\begin{itemize}
\item \textbf{Semantic adequacy}: Does the text correctly represent the meaning in the data?
\item \textbf{Grammaticality}: Is the text grammatical (no spelling or grammatical errors)?
\item \textbf{Fluency}: Does the text sound natural?
\end{itemize}

\cite{castro-ferreira-etal-2020-2020} provides human judgments for the output of 16 NLG systems from \wnlg\ Challenge 2020. Each model was evaluated on a sample of 178 texts yielding a total of 2,848 generated texts annotated with human judgments for the following five criteria.

\begin{itemize}
\item \textbf{Data Coverage}: Does the text include descriptions of \textit{all} predicates in the input?
\item \textbf{Relevance}: Does the text describe \textit{only} triples present in the graph?
\item \textbf{Correctness}: For graph predicates, does the text correctly describe their arguments?
\item \textbf{Text Structure}: Is the text grammatical, well-structured, written in acceptable English?
\item \textbf{Fluency}: Does the text progress naturally and form a coherent, easy-to-understand whole?
\end{itemize}

We train on the 2017 \textit{semantic adequacy} metric. To assess how well our similarity metric reflects human judgments of similarity between an RDF graph and a Natural Language Text, we compute correlations between our system's scores and the 2020 human judgments of semantic adequacy, namely \textit{data coverage}, \textit{relevance}, and \textit{correctness}\footnote{We train on \wnlg{}-2017 and evaluate on \wnlg{}-2020 as semantic adequacy is a more global criterion encompassing coverage, relevance and correctness while the reverse is not true.}.

\subsection{Fine-tuning Procedure}

\begin{figure*}[!h]
    \centering
    \includegraphics[width=0.9\linewidth]{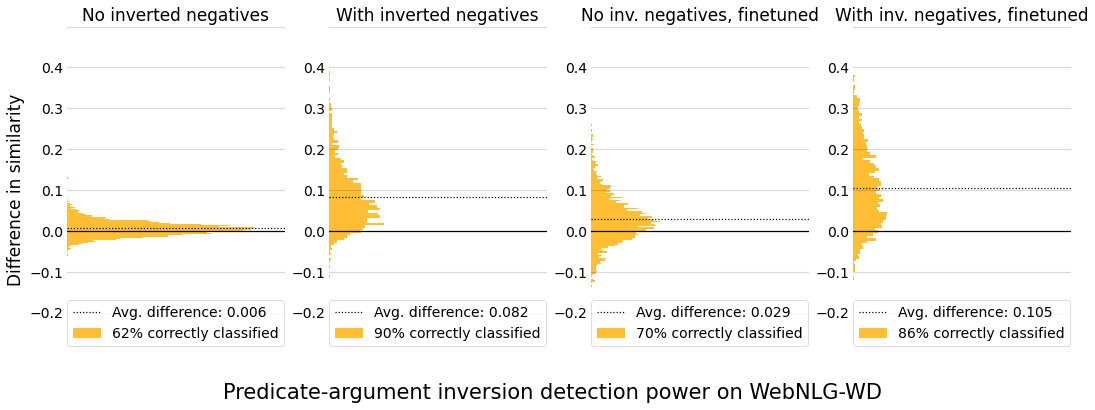}
    \caption{\textbf{Difference in similarity between correct and corrupted graph-text pairs.} On the left,  \textcolor{allhard}{\texttt{all\_datasets\_hard\_negatives}} and \textcolor{allinvhard}{\texttt{all\_datasets\_hardinv\_negatives}} just after pre-training, and on the right, both models after fine-tuning and ensembling on \wnlg{}-2017. The system we used as a final metric is the last plot on the right. Models that have seen inverted negatives at pre-training identify correct and corrupted pairs better.}
    \label{fig:invsim} 
\end{figure*}

\paragraph{Bi- and Cross-encoder ensembling} We can fine-tune our pre-trained model as a \textit{cross-encoder}, where there is only one instance of the model, which can attend to both items simultaneously and feed into a linear layer, rather than a \textit{bi-encoder} as previously, where two instances of the model embed the two items separately and the dot product or cosine distance serves as the output. The cross-attention feature allows for higher performance at the cost of making retrieval expensive as all $n^2$ distances must be computed separately \cite{Polyencoders}. However, bi- and cross-encoders perform well on different data points. The scores they give \wnlg-2020 candidates have surprisingly low Pearson correlation, 0.66. This makes them good candidates for ensembling, and indeed, taking the mean of the bi- and cross-encoder scores yields higher correlations with all human judgments. Both architectures and the ensembling method are represented in diagram \ref{fig:ensembling_archs}.

\begin{figure*}[!ht]
\centering
    \includegraphics[width=0.9\linewidth,scale=0.1]{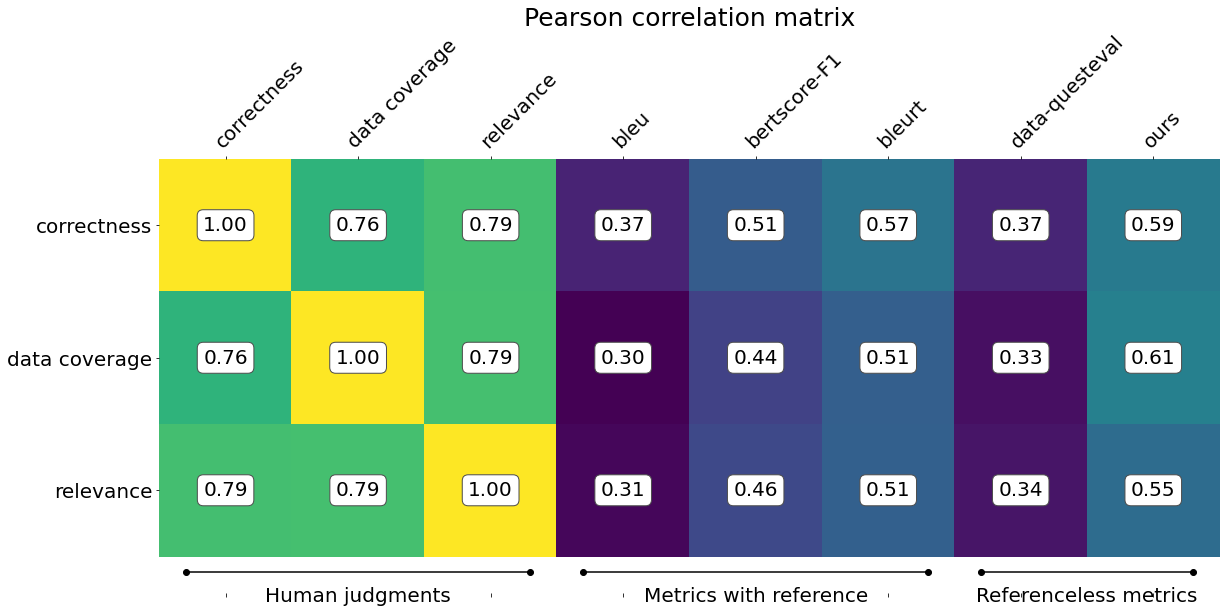}
    \caption{\textbf{Pearson correlation between automatic metrics and human judgments.} Lighter and higher is better. \mo{} outperforms the other referenceless metric and matches BLEURT, which requires a reference.}\label{fig:correlation}
\end{figure*}


\paragraph{Robustness to inversion} Transformer-based models can sometimes behave as advanced bag-of-word models \citep{wordorder}, which would not see a difference if the subject and object are reversed in a triple. In order to examine the robustness, we create an adversarial dataset from all the 1-triple graphs in \wnlg\ 2020 with non-symmetrical\footnote{Manually defined. The list is in appendix \ref{sec:symmetrical}.} relationships. In this dataset, for each text, there is a pair with the correct triple and a pair in which the triple's predicate arguments (subject and object) have been inverted e.g., \fact{André the Giant}{larger than}{Samuel Beckett} vs. \fact{Samuel Beckett}{larger than}{André the Giant}. This dataset (\wnlginv ) consists of 2793 $(g,t),$ and  $(g\_{inv}, t)$ pairs where $(g,t)$ is a graph of size one with a non-symmetrical relationship in \wnlgwd, $t$ is the corresponding text and $g\_{inv}$ is the corrupted triple. 

We report the difference $sim(g,t) - sim(g_{inv},t)$ in the similarity between text and correct graph on the one hand and text and corrupted graph on the other in Figure \ref{fig:invsim}. The higher, the better the model is at recognizing predicate inversion.  \textcolor{allhard}{\texttt{all\_datasets\_hard\_negatives}}, the retrieval model presented in Section~\ref{subsec:model}, does not do well at this task, with 38\% of the inverted triplets estimated more similar to the text than the original ones. (After fine-tuning on \wnlg{}-2017 judgments, 30\%)

In order to make our models robust to inversion, at pre-training time, we add inverted negatives to the mix of artificial negatives in the batches: confounding graphs where a random triplet has been inverted. The resulting model,  \textcolor{allinvhard}{\texttt{all\_datasets\_hardinv\_negatives}} has the same retrieval accuracy but gains inversion detection abilities. This ability is conserved through fine-tuning, as Figure~\ref{fig:invsim} shows: only 14\% of triplets are misclassified.

\paragraph{The final system we choose as a metric}is the ensemble of a bi- and cross-encoder pre-trained on the concatenation of \kelm{}, \TeKGen{} and \trex{} with our two types of data augmentation, then fine-tuned on \wnlg{}-2017 human judgments. We call it \mo{}, for Ensembled Representations for Evaluation of DAta-to-Text.

\subsection{Comparison with other Evaluation Metrics}

\begin{figure}[]
\centering
    \includegraphics[width=0.9\linewidth]{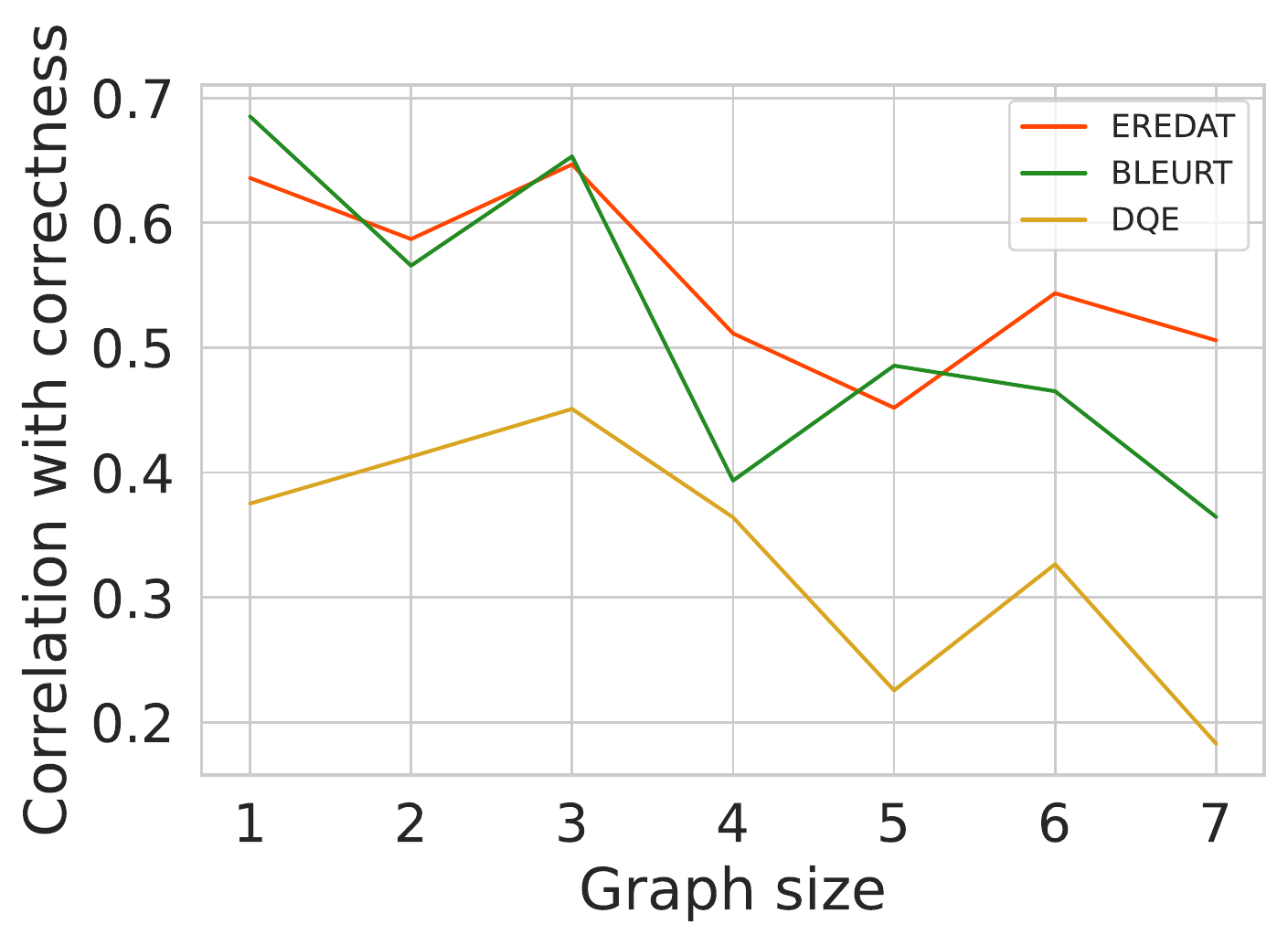}
    \caption{\textbf{Correlation with human judgment by graph size} for \mo{}, DQE and BLEURT. Our metric is more robust than BLEURT to longer graphs, and generally much more correlated than DQE, the existing referenceless metric.}\label{fig:graph_size}
\end{figure}

Correlations with human judgments are shown in Figure~ \ref{fig:correlation} for a variety of automated evaluation metrics: three metrics that require a reference (BLEU, BERTscore-F1, and BLEURT, the previous state of the art) and two referenceless metrics (Data-QuestEval and \mo{}). Our metric is the best correlated with all human judgment categories, even including metrics with references. As shown in \ref{fig:graph_size}, this advantage is mostly explainable by \mo{}'s improved robustness to longer, more complex graphs, which tend to degrade correlation with human judgment. Scatter plots of the underlying distributions are given in appendix \ref{sec:scatter}.

As human references are rarely available and costly to produce, and \mo{} attains higher correlation with human judgments without relying on them, it is the most practical choice to evaluate data-to-text generation. In this case, it was not fine-tuned to the same kind of data it was applied to, showing it generalizes to new datasets. If one has a specific dataset or task in mind, even better performance could be attained by training on a set of problem-specific human judgments.

\section{Conclusion}

We presented an architecture and pre-training strategy to measure the similarity between RDF graphs and English texts, introducing novel data augmentation strategies made possible by the RDF structure. Specifically, we introduced a bi-encoder retrieval model trained on unlabeled RDF-text data which achieves high retrieval accuracy on both parallel and real-life, less well aligned datasets. Building from this pre-trained model, we further provided a novel evaluation metric for RDF-to-text generation models which matches state-of-the art metrics in terms of correlation with human judgments of semantic adequacy without needing costly human-written references. This metric can also be used to filter existing  text/RDF datasets. 

\bibliographystyle{acl_natbib}
\bibliography{anthology, custom}

\newpage

\appendix

\section{Dataset statistics}
\label{sec:retrievaldata}

\begin{table}[!ht]
\small
\centering
\begin{tabular}{lccccc}
\toprule
 & \# (t,g) & \# P & \# E
\\\midrule
\TeKGen & 6,310,061 & 1041 & 3,939,696
\\
\trex & 6,000,336 & 675 & 3,188,309
\\
\kelm & 15,616,551 & 261405 & 5,073,603
\\\midrule
\wnlgdb & 13,212 & 372 & 3210
\\
\wnlgwd & 10,384 & 188 & 2783
\\
\wchunks & 30,000 & 468 & 20,318
\\\bottomrule
\end{tabular}
\caption{\textbf{Training and test data for retrieval}. \# (t,g): Number of graph-text pairs, \# T: Number of texts, \# G: Number of graphs, \# P: Number of distinct properties, \# E: Number of distinct entities.}
\end{table}

\section{Impact of Batch Size}
\label{sec:batchsize}

\begin{figure*}[!ht]
    \centering
    \includegraphics[width=\linewidth]{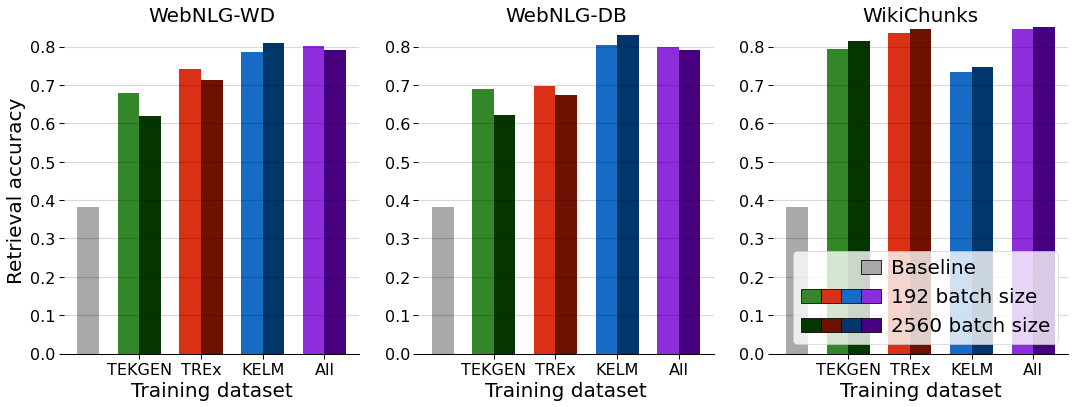}
    \caption{\textbf{Small vs. Large Batch Size.} Large batch sizes help a little on data with lower alignement quality (\wchunks ). Overall, the improvement is inconsistent.}
    \label{fig:batchsize}
\end{figure*}

\section{Scatter Plot Comparison of BLEURT and \mo{}}
\label{sec:scatter}

\begin{figure*}[!ht]
    \centering
    \includegraphics[width=\linewidth]{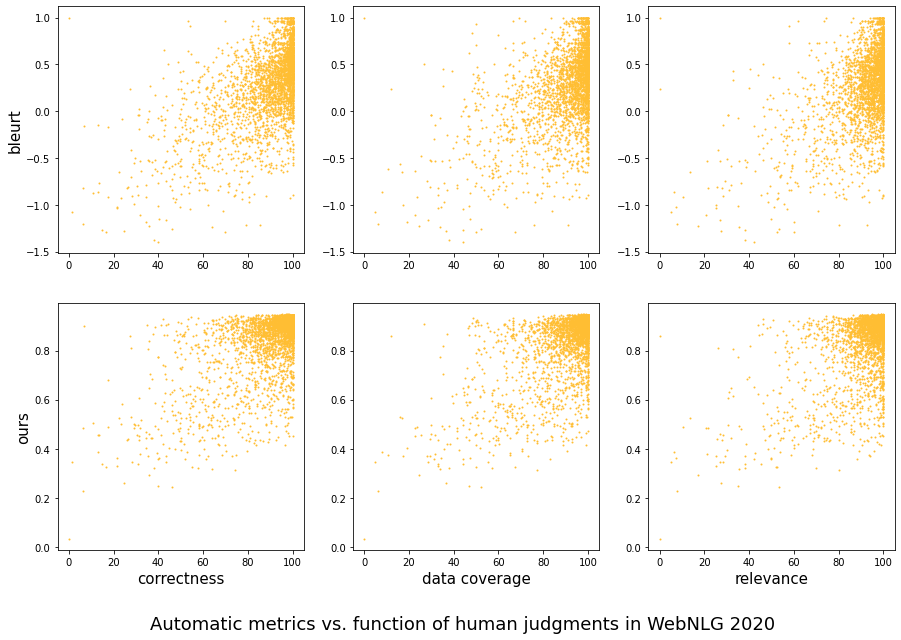}
    \caption{\textbf{Human judgment and automated evaluation values for every point in \wnlg{} 2020.}}
    \label{fig:bleurt_vs_ours}
\end{figure*}

\newpage

\section{Symmetrical Relationships in WebNLG}
\label{sec:symmetrical}

We manually inspected all relationships in \wnlg{} and deemed the following to be symmetrical in nature:

"taxon synonym",
"partner in business or sport",
"opposite of",
"partially coincident with",
"physically interacts with",
"partner",
"relative",
"related category",
"connects with",
"twinned administrative body",
"different from",
"said to be the same as",
"sibling",
"adjacent station",
"shares border with"

\end{document}